# Source-free unsupervised domain adaptation for cross-modality abdominal multi-organ segmentation

Jin Hong[1,2,#], Yu-Dong Zhang[3,#], Weitian Chen[1,2,*]

1. CU Lab of AI in Radiology (CLAIR), the Chinese University of Hong Kong, Shatin, Hong Kong SAR, China
2. Department of Imaging and Interventional Radiology, the Chinese University of Hong Kong, Shatin, Hong Kong SAR, China
3. School of Informatics, University of Leicester, Leicester LE1 7RH, UK

E-mail: hongjin_91@163.com; yudongzhang@ieee.org; wtchen@cuhk.edu.hk;

# Those two authors contributed equally to this work, and they should be considered as co-first authors.
* Correspondence should be addressed to Weitian Chen

**Abstract:** Domain adaptation is crucial for transferring the knowledge from the source labeled CT dataset to the target unlabeled MR dataset in abdominal multi-organ segmentation. Meanwhile, it is highly desirable to avoid the high annotation cost related to the target dataset and protect the source dataset privacy. Therefore, we propose an effective source-free unsupervised domain adaptation method for cross-modality abdominal multi-organ segmentation without source dataset access. The proposed framework comprises two stages. In the first stage, the feature map statistics-guided model adaptation combined with entropy minimization is developed to help the top segmentation network reliably segment the target images. The pseudo-labels output from the top segmentation network are used to guide the style compensation network to generate source-like images. The pseudo-labels output from the middle segmentation network is used to supervise the learning progress of the desired model (bottom segmentation network). In the second stage, the circular learning and pixel-adaptive mask refinement are used to further improve the desired model performance. With this approach, we achieved satisfactory abdominal multi-organ segmentation performance, outperforming the existing state-of-the-art domain adaptation methods. The proposed approach can be easily extended to situations in which target annotation data exist. With only one labeled MR volume, the performance can be levelled with that of supervised learning. Furthermore, the proposed approach is proven to be effective for source-free unsupervised domain adaptation in reverse direction.

## 1. Introduction

Over the past few decades, deep learning has been successful in the medical image analysis field, when large-scale densely labeled datasets are available (Çiçek et al., 2016; Milletari et al., 2016; Ronneberger et al., 2015). However, this condition is often violated in real-world clinical scenarios owning to the high annotation cost of medical images. Transferring knowledge from models well trained on labeled source datasets to unlabeled target datasets is an intuitive approach to address this issue. However, this tends to be hindered by the well-known domain-shift problem when the data distributions in the source and target domains are inconsistent (Kouw and Loog, 2019). This often occurs in medical imaging because medical images are acquired using different modalities, scanners, protocols, sites, and populations that can have different characteristics. For instance, images acquired using magnetic resonance imaging (MRI) and computed tomography (CT) can be significantly different because of the different physical principles of these imaging modalities. Such cross-modality domain shifts can significantly degrade the performance of deep neural networks in medical imaging if not properly addressed (Chen et al., 2020; Novosad et al., 2019; Yang et al., 2019).

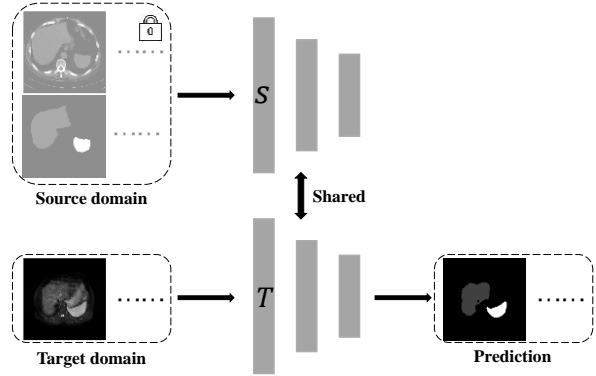

Figure 1 Illustration of source-free unsupervised domain adaptation for segmentation

Unsupervised domain adaptation (Bousmalis et al., 2017; Chang et al., 2019; French et al., 2017; Ganin et al., 2016; Hoffman et al., 2018; Hoffman et al., 2016; Tzeng et al., 2017; Vu et al., 2019; Wu et al., 2018; Zhu et al., 2017; Zou et al., 2018) was developed to address the domain shift problem. It generalized the models trained on the labeled source domain to the unlabeled target domain to reduce expensive data labeling costs, especially labeling medical images (Luo et al., 2020; Luo et al., 2021). All these methods require the availability of labeled source datasets and well-trained models to learn the source knowledge during domain adaptation training. However, nowadays, the datasets are often private with limited access owning to commercial or regulatory reasons in fields such as commerce, finance, and healthcare. Recently, governments, including the European Union (EU), have announced various data-protection frameworks. One example is the General Data Protection Regulation (GDPR) that highlights the security issues of data transmission (Liang et al., 2021). Such regulations prohibit existing unsupervised domain adaptation methods from accessing source data because they violate the data privacy policy. Source-free unsupervised domain adaptation methods (Kim et al., 2021; Kundu et al., 2020; Li et al., 2020; Liang et al., 2019; Liang et al., 2020; Sahoo et al., 2020) can be used to address this challenge, as illustrated in Figure 1. Most of these methods focus on image-classification tasks and very few focus on image segmentation. Note that image segmentation is a pixel-level task where each pixel is associated with a semantic label; thus, it is a more complex issue than image classification (Liu et al., 2021).

Accurate segmentation of organs in abdominal images acquired using CT or MRI plays an important role in many clinical scenarios such as surgery planning, computer-assisted diagnosis, visual augmentation, and image-guided interventions (Summers, 2016). In clinical practice, CT and MRI are often used to provide complementary diagnostic information. It is valuable to provide segmentations of the abdominal organs using both modalities for proper the clinical management of abdominal diseases. Currently, CT images are more commonly used than MR images because of lower costs and wider availability. Accordingly, it is desirable to achieve source-free unsupervised domain adaptation to transfer the learned knowledge from the source model, well trained on labeled CT images, to the unlabeled target MR images.

In this study, we propose a two-stage source-free unsupervised domain-adaptation framework for cross-modality abdominal multi-organ segmentations. Our proposed method only requires a well-trained source model and unlabeled target dataset and does not require access to the source data. The proposed method is illustrated in Figure 2. Our experimental results highlight that this approach achieves satisfactory performance for adaptation of a labeled CT dataset to an unlabeled MR dataset on abdominal organ segmentations including the liver, kidneys, and spleen with Dice similarity coefficients of 0.884, 0.891, 0.864, and 0.911, respectively. In addition, the proposed approach also achieves satisfactory performance for adaptation in the reverse direction. The main contributions of this study are summarized as follows:

(i) We firstly extend the no prior-aware source-free unsupervised domain adaptation method to bidirectional cross-modality medical image segmentation with unpaired CT and MR images.

(ii) The proposed source-free unsupervised domain adaptation framework organically combines feature alignment, image alignment, and self-learning methods. Consequently, it can achieve a high performance, outperforming the existing state-of-the-art domain adaptation methods.

(iii) We develop a simple but effective source-free feature map statistics-guided model adaptation method to achieve reliable segmentations of target images.

(iv) We propose an image transformation method called style compensation guided by pseudo-labels to transform the target images to source-like images and retain the target-like contrast between organs.

(v) We propose an extension module that can be easily embedded into the proposed unsupervised domain adaptation framework to utilize labeled target data if they exist. The proposed method with this extension module can achieve a performance close to that of fully supervised learning (upper bound) by using only a single labeled target volume.

## 2. Related work

### 2.1. Unsupervised domain adaptation

Unsupervised domain adaptation is a typical example of transfer learning (Pan and Yang, 2009). It aims to exploit the knowledge depicted in labeled datasets to achieve a discriminative model for different but related unlabeled datasets. Several unsupervised domain adaptation methods have been developed to reduce cross-domain discrepancies, including feature-level alignment (Chen et al., 2017; Dou et al., 2018; Ganin et al., 2016; Hoffman et al., 2016; Kamnitsas et al., 2017; Sun and Saenko, 2016; Tzeng et al., 2017; Tzeng et al., 2014; Vu et al., 2019), pixel-level alignment (Bousmalis et al., 2017; Chang et al., 2019; Yang et al., 2019; Zhu et al., 2017), self-learning (French et al., 2017; Novosad et al., 2019; Perone et al., 2019; Zou et al., 2018), and their mixtures (Chen et al., 2020; Hoffman et al., 2018; Hong et al., 2022; Wu et al., 2018).

Most unsupervised domain adaptation methods focus on aligning the feature distributions extracted from the unlabeled source and labeled target datasets. Currently, generative adversarial networks (GANs) (Goodfellow et al., 2020) and their variants (Arjovsky et al., 2017) are often used to reduce the domain shift and implicitly align feature distributions. Hoffman et al. (2016) applied adversarial learning for the first time to a fully convolutional network to implicitly learn the mapping of features between the source and target domains. Chen et al. (2017) conducted a similar study to align global and class-wise domains using adversarial learning. Ganin et al. (2016) used adversarial learning to learn domain-invariant features using weights shared by two networks. Tzeng et al. (2017) developed a flexible unsupervised domain adaptation framework named adversarial discriminative domain adaptation (ADDA). Vu et al. (2019) proposed an unsupervised domain adaptation framework for semantic segmentation based on entropy. For medical image analysis, Kamnitsas et al. (2017) utilized an adversarial learning-based method to train a network to be invariant under different inputs for brain lesion segmentation in MR images. Dou et al. (2018) transferred the knowledge learned from labeled MR images to unlabeled CT images for cardiac segmentation.

Another stream of studies focuses on pixel-level alignment to transform the "style" of the target images to that of the source images, or vice versa. Bousmalis et al. (2017) converted source images with appearances similar to those of the target images using a GAN-based method. Zhu et al. (2017) developed the so-called cycle-GAN to generate target images from unpaired source images for pixel-level adaptation. Chang et al. (2019) proposed a domain-invariant structure extraction (DISE) framework to achieve domain adaptation at the pixel level by decomposing images into domain-invariant structures and domain-specific textures. Similarly, Yang et al. (2019) disentangled medical images into domain-invariant content and domain-specific styles using adversarial learning to achieve domain adaptation of liver images at the pixel level between CT and MRI scans.

Some studies implemented unsupervised domain adaptation via self-learning, aiming to train the current network with the pseudo-labels generated by the previous state of the model or ensembled model, which is usually used in semi-supervised methods (Laine and Aila, 2016; Tarvainen and Valpola, 2017). French et al. (2017) proposed a self-ensembling method for visual domain adaptation. Zou et al. (2018) proposed an unsupervised domain adaptation for semantic segmentation using a self-learning method that combines class balancing and spatial prior information. Perone et al. (2019) utilized a self-ensembling approach based on the mean teacher model to achieve unsupervised domain adaptation for spinal cord segmentation using MR images. Novosad et al. (2019) applied self-ensembling to improve their unsupervised domain adaptation framework for neuroanatomical segmentation.

Several frameworks combine the three aforementioned approaches for unsupervised domain adaptation. Hoffman et al. (2018), Wu et al. (2018), and Chen et al. (2020) adapted the target domain to the source domain at both pixel- and feature-level using cycle-consistent adversarial domain adaptation (CyCADA), dual channel-wise alignment networks (DCAN), and synergistic image and feature alignment (SIFA) method, respectively. Hong et al. (2022) developed a novel unsupervised domain adaptation framework that combines adversarial learning and self-learning for cross-modality liver segmentation.

**2.2. Source-free unsupervised domain adaptation**

All the aforementioned methods require source datasets to be available during domain adaptation. In other words, these methods are inapplicable if the source datasets are unavailable. Several source-free unsupervised domain adaptation methods have been developed to address this issue in image classification and segmentation.

Liang et al. (2019) proposed an effective method based on subspace discovery to achieve a source-free unsupervised domain adaptation for image classification. Kundu et al. (2020) developed a two-stage framework for source-free unsupervised domain adaptation. In this approach, the source model is first trained to reject out-of-source distribution samples. Second, a unified adaptation method is designed to operate across a wide range of category gaps. Li et al. (2020) proposed a collaborative-class conditional generative adversarial network. In Li's approach, the prediction model can offer more accurate guidance for the generator using the generated target-style data. Liang et al. (2020) developed a generic representation-learning framework called Source HypOthesis Transfer (SHOT). Here, the classifier module (hypothesis) was frozen in the source model. The information maximization and self-supervised pseudo-labeling were used to learn target-specific features and implicitly align target representations to the source hypothesis. Sahoo et al. (2020) proposed a method to adapt the source classifier to the target domain using a source classifier and unlabeled target samples. Kim et al. (2021) proposed an approach that utilized a well-trained source model that progressively learned a target model via self-learning.

These source-free unsupervised domain adaptation methods have been proposed to address image classification tasks. Image classification is an image-level task that associates a label with the entire image. Contrarily, image segmentation is a pixel-level task that assigns each pixel to a category. Obviously, the latter task is more complex because it must adapt the source model to accurately classify each pixel in the target domain. Inspired by data-free knowledge distillation (Chen et al., 2019b; Fang et al., 2019; Yin et al., 2020), Liu et al. (2021) developed a two-stage source-free unsupervised domain adaptation framework for segmentation of natural images. This method contains a dual-attention distillation (DAD) mechanism for knowledge transfer and an entropy-based intra-domain patch-level self-supervision module (IPSM) for model adaptation. Wang et al. (2020) implemented source-free unsupervised domain adaptation on semantic segmentation using test entropy minimization (tent). The tent adapted the model during testing and minimized the entropy of its predictions by modulating its features. These feature modulations were updated by estimating the normalization statistics (mean and standard deviation) using the target data and optimizing the two transformation parameters.

In the medical image analysis field, Chen et al. (2021) proposed a denoised pseudo-labeling strategy for source-free unsupervised domain adaptation in fundus image segmentation. Two complementary pixel-level (uncertainty estimation) and class-level (prototype estimation) denoising methods were used to reduce the noisy pseudo-labels achieved by inputting the target data into the source model and determine the reliable labels, so that the target model could be guided to implement domain adaptation. Bateson et al. (2021) developed a source-free domain adaptation method with a constrained formulation that embedded domain-invariant prior knowledge on the regions of interest for medical image segmentation. The target model is initialized using the weights of the source model and optimized by combining entropy minimization and class ratio of unlabeled target data acquired from anatomical knowledge. In this study, we developed a no-prior-aware source-free unsupervised domain adaptation method for cross-modality multi-organ segmentation. A feature map statistics-guided model adaptation method was proposed to reliably segment the target image. A style compensation network was developed to significantly improve the segmentation accuracy of the left kidney and spleen. Pixel-adaptive mask refinement was employed to eliminate unreasonable isolated pixels, while circular learning was used to further improve the performance.

## 3. Method

### 3.1. The source-free unsupervised domain adaptation framework

In this study, we define the source-free unsupervised domain adaptation task as transferring the knowledge present in model ($S$) trained on the source domain ($X_s, Y_s$) with labeled CT images to the target domain ($X_t$) with unlabeled MR images. We propose a novel source-free unsupervised domain adaptation framework for cross-modality abdominal multi-organ segmentation without accessing the source domain datasets. Our proposed framework includes three segmentation networks (U1 (representing the combination of U1-1 and U1-2 in this study), U2, and U3) and one style compensation network (SC) as shown in Figure 2. The network U3 represents the desired model that is used to segment abdominal organs in MR images, and the other networks are designed to assist in achieving this goal. The backend weights of the source model $S$ are shared with U1-2, and all weights of the source model are shared with U2. Those weights were frozen during training (indicated by the gray color in Figure 2). After domain adaptation (training), abdominal organ segmentation can be achieved by inputting the MR images into the desired model U3 with a single forward pass.

As shown in Figure 2, the training process can be divided into two stages. In the first stage ($Epoch < T$), the entire framework is optimized in a one-way manner from the top to the bottom network (U1-1→SC→U3). U1-1 learns the knowledge guided by the feature map statistics stored in U1-2 and entropy minimization of the predictions. The pseudo-label $y_{1,t}$ generated by U1-2 is used to supervise the SC network to obtain the style compensation coefficient $p_s$ to transform the MR image $x_t$ to the CT-style image $x_{t\to s}$, which can be accurately segmented by the source model U2. Similarly, the pseudo-label $y_{2,t}$ generated by U2 is used to supervise the desired model U3 to accurately segment abdominal organs directly from the MR image. In the second stage ($Epoch \geq T$), the entire framework is optimized in a circular manner (U1-1→SC→U3→U1-1). While keeping the original optimization mechanism unchanged, pixel-adaptive mask refinement is introduced to provide self-supervision to further improve the segmentation accuracy of U3. The pseudo-label $y_{3,t}$ generated by U3 is added to supervise U1-1 as a new constraint to further improve its performance, forming a positive feedback optimization mechanism. Note that the extension module is designed to increase the flexibility of the proposed method. This is optional and not required by our proposed source-free unsupervised domain-adaptation framework. More details can be found in section 4.6.

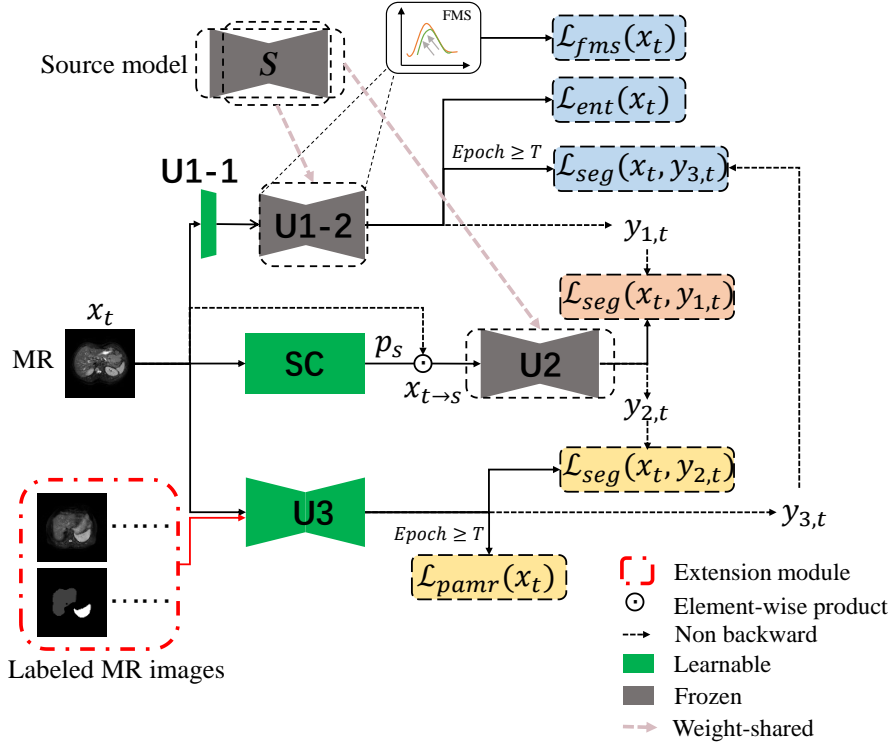

Figure 2 (Best viewed in color) Overview of the proposed source-free unsupervised domain adaptation framework. The gray areas denote the weights in the networks that are frozen during training, while the weights in the green areas are updatable. The dotted red box refers to an optional extension module that can be used to handle the addition of labeled MR images to train the desired model U3 in a fully-supervised learning manner.

## 3.2. Segmentation network

In the proposed source-free unsupervised domain adaptation framework, all segmentation networks U1, U2, and U3 adopt the structure of the classical semantic segmentation network U-Net (Ronneberger et al., 2015), which is most commonly used in the medical image analysis field. The structure of the proposed segmentation network is presented in Figure 3. Note that it contains nine convolution blocks (blue triangles), one convolution layer (brick red triangles), four max-pooling layers (yellow triangles), four up-sampling blocks (purple triangles), and four skip-connection operations (gray triangles). Each convolution block consists of two stacked convolution layers, each of which is followed by a batch normalization layer (Ioffe and Szegedy, 2015) and rectified linear unit (ReLU) layer (Nair and Hinton, 2010). Each up-sampling block contains an up-sampling layer, convolution layer, batch normalization layer, and ReLU layer. The convolution block attempts to extract as many task-related features as possible, while the max-pooling layer is used as a down-sampling operation to select more critical features. An up-sampling block is employed to enlarge the feature maps and smooth them out. The skip connection operation is used to concatenate high-level feature maps and their corresponding low-level counterparts to achieve more fine-grained segmentations, which is the key difference between U-Net and traditional convolutional neural networks (CNNs).

Given a labeled dataset containing $N$ samples $(X, Y) = \{(x^1, y^1), \cdots, (x^N, y^N)\}$, $x^i$ denotes the abdominal MR image and $y^i$ denotes the anatomical structure of the organs. After image $x^i$ passes through the segmentation network, it outputs a prediction map $o^i$ that is supposed to approximate the ground truth $y^i$. In this study, we utilize the Dice coefficient as the error (distance) between $o^i$ and $y^i$ to optimize the segmentation network. Note that the Dice coefficient was proposed to avoid the predictions being strongly biased towards the background (Milletari et al., 2016). Thus, segmentation loss can be defined as follows:

$$\mathcal{L}_{seg}(X, Y) = \frac{\sum_{i=1}^{N} 2 y^i o^i}{\sum_{i=1}^{N} y^i y^i + \sum_{i=1}^{N} o^i o^i} \quad (1)$$

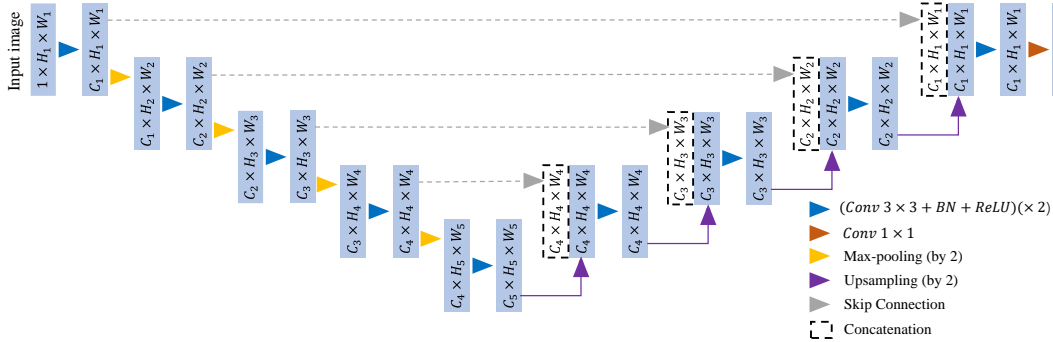

Figure 3 (Best viewed in color) Structure of the proposed segmentation networks. The size of feature maps is represented by $C \times H \times W$. $C$, $H$ and $W$ denote the depth, height and width respectively.

## 3.3. Feature map statistics-guided model adaptation

Yin et al. (2020) developed a data-free knowledge distillation framework (Figure 4a) to synthesize images from a pre-trained model without using any additional information from the training dataset. As shown in Figure 4a, keeping the teacher (pre-trained model) fixed, the input random noise is updated to synthesize class-conditional images supervised under the cross-entropy loss provided by the target class and regularized by the feature map statistics information (mean and variance) stored in the teacher model. Meanwhile, the Jensen-Shannon divergence between the teacher and student model logits are maximized to improve the diversity of the synthesized images.

Accordingly, Liu et al. (2021) first introduced feature map statistics regularization in their proposed framework to transfer knowledge from the source model to the target model without accessing the source dataset for semantic segmentation. Here, we extract the module (feature map statistics-guided knowledge transfer) from their framework as illustrated in Figure 4b. Unlike the data-free knowledge distillation framework (Figure 4a), the feature map statistics-guided knowledge transfer method introduces a generator for synthesizing fake images instead of updating the random noise directly. Keeping the source model fixed, the generator is supervised under the feature map statistical information stored in the source model. The generator aims to synthesize images whose distributions are similar to those of the source images. Here, maximizing the mean absolute error between the source and target model predictions improves the diversity in the synthesized images. To perform domain adaptation, Liu et al. (2021) introduced another module for model adaptation.

In the feature map statistics-guided knowledge transfer method, the generator easily falls into mode collapse and generated fake images have almost no specific practical meaning. To avoid mode collapse, we develop a feature map statistics-guided model adaptation method that abandons the generator and uses a noisy image as the input. This approach is easier to implement and can directly obtain an adapted model as shown in Figure 4c. We assume that the discrepancy between the target and source feature distributions can be eliminated in shallow layers. The first convolution block in the source model is removed and an encoder is designed to output source-like feature maps with target images under the supervision of feature map statistics stored in the remaining part of the fixed source model. Using this approach, it is easy to achieve model adaptation without accessing the source data and combination of the encoder after training. The remaining part of the fixed source model is the desired target model that can be used to reliably segment the target images.

Feature map statistics regularization minimizes the distance between the feature map statistics of the source image $x_s$ and target image $x_t$. It assumes that feature maps at all levels obey a Gaussian distribution across batches; therefore, the feature map statistics can be described as mean $\mu$ and variance $\sigma^2$. Therefore, regularization is defined as follows:

$$\mathcal{L}_{fms}(x_t) = \sum_l \|\mu_l(x_t) - \mu_l(x_s)\|_2 + \sum_l \|\sigma_l^2(x_t) - \sigma_l^2(x_s)\|_2 \qquad (2)$$

where $\mu_l(x)$ and $\sigma_l^2(x)$ are the batch-wise mean and variance of the feature maps in the $l$th convolution layer, respectively. $\|\cdot\|_2$ denotes $l_2$ norm calculations.

Note that the running average statistics (channel-wise means and variances) are stored in the batch normalization layers when the source models are trained using the source images. The batch normalization layer is widely used in deep neural networks, including U-Net, to alleviate the covariate shifts (Ioffe and Szegedy, 2015). Therefore, the feature map statistics $\mu$ and variance $\sigma^2$ of the source images can be obtained from the source model without accessing the source dataset.

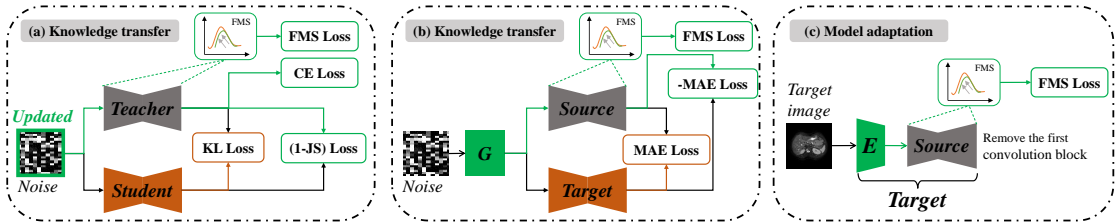

Figure 4 (Best viewed in color) Comparison of our feature map statistics-guided model adaptation with other two data-free knowledge transfer methods. (a) Data-free knowledge distillation framework (Yin et al., 2020); (b) Feature map statistics-guided knowledge transfer (Liu et al., 2021); (c) Feature map statistics-guided model adaptation. FMS, CE, KL, JS, and MAE denote the feature map statistics, cross entropy, Kullback–Leibler divergence, Jensen-Shannon divergence, and mean absolute error respectively. G and E denote the generator and encoder respectively.

**3.4. Entropy minimization**

Source models tend to produce overconfident (low-entropy) predictions on source-like images and underconfident (high-entropy) predictions on target-like images (Vu et al., 2019). Consequently, entropy minimization is utilized as an important constraint to improve the prediction results in many semi-supervised learning, domain adaptation, and clustering methods (Grandvalet and Bengio, 2005; Jain et al., 2017, 2018; Long et al., 2016; Springenberg, 2015). In our approach, we introduced entropy minimization loss as self-supervision to directly minimize the prediction entropy in domain adaptation. Given an unlabeled dataset $X$ containing $N$ images of size $H \times W$ and a pixel $o^{n,h,w}$ from the corresponding $N$ prediction maps, the entropy loss $\mathcal{L}_{ent}(X)$ for our tasks is defined as:

$$\mathcal{L}_{ent}(X) = \frac{\sum_{n=1}^{N}\sum_{h=1}^{H}\sum_{w=1}^{W}(-o^{n,h,w}\log(o^{n,h,w}))}{N*H*W} \quad (3)$$

**3.5. Style compensation**

Assuming that images can be decomposed into domain-invariant content and domain-specific style subspaces by disentangling the latent space, two types of approaches have been proposed for image-to-image translation (or domain adaptation) (Chang et al., 2019; Huang et al., 2018; Mathieu et al., 2016; Siddharth et al., 2017; Yang et al., 2019). (i) Transforming the source images to appear similar to the target images (or vice versa) by replacing the style representations of the source images with those of the target images, as shown in Figure 5a. (ii) Removing the style representations of both the source and target images and retaining their content representations, as shown in Figure 5b. These methods are performed given that both the source and target images are available; thus, are inapplicable to our tasks.

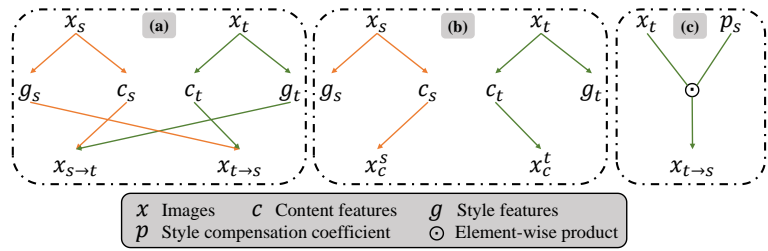

Figure 5 (Best viewed in color) Illustration of cross-domain image translation methods

Here, we propose a third method called style compensation (see Figure 5c) to translate the target images into source-like images without accessing the source images. This method is embedded in the proposed source-free unsupervised domain-adaptation framework. As demonstrated in Figure 5c, the source-like image $x_{t \to s}$ can be formed by multiplying image $x_t$ by the style compensation coefficient $p_s$ in an element-wise manner (note that both $x_t$ and $p_s$ have a size $1 \times H \times W$):

$$x_{t \to s} = x_t \cdot p_s \quad (4)$$

To obtain the style compensation coefficient $p_s$, we designed the network presented in Figure 6. It consists of several stacked convolution layers, followed by batch normalization and activation-function layers. It abandons pooling layers and maintains the height and width of the feature maps the same as the input to preserve the fine structures contained in the style compensation coefficient. As shown in Figure 2, the style compensation network is optimized by the segmentation loss $\mathcal{L}_{seg}(x_t, y_{1,t})$ using the pseudo-label $y_{1,t}$ generated by the output of U1-2.

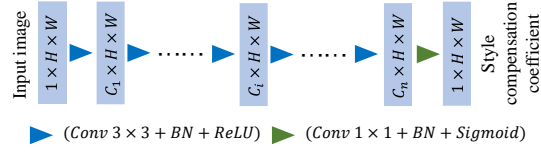

Figure 6 (Best viewed in color) Illustration of style compensation network

### 3.6. Pixel-adaptive mask refinement

Araslanov and Roth (2020) defined local consistency as an indicator of whether neighboring pixels with similar appearances in a semantic mask fall into the same category. They proposed the so-called pixel-adaptive mask refinement to improve local consistency. As a mask refinement approach, pixel-adaptive mask refinement can provide a self-supervision for semantic segmentation models to further improve performance, such as GrabCut (Rother et al., 2004) and dense CRFs (Krähenbühl and Koltun, 2011).

Given a pixel-level semantic mask $o_{:,:,:} \in (0,1)^{C \times H \times W}$ ("$C$" represents $C$ categories containing $C-1$ kinds of objects and one background) and an input image $x$, pixel-adaptive mask refinement aims to update each pixel label $o_{:,i,j}$ iteratively according to the convex combination of its neighbors $\mathcal{N}(i,j)$. At the $t^{\text{th}}$ iteration, the refined pixel label $o_{:,i,j}$ is expressed as follows:

$$o_{:,i,j}^t = \sum_{(m,n) \in \mathcal{N}(i,j)} \alpha_{i,j,m,n} \cdot o_{:,m,n}^{t-1} \tag{5}$$

$$\alpha_{i,j,m,n} = \frac{e^{\bar{k}(x_{i,j}, x_{m,n})}}{\sum_{(p,q) \in \mathcal{N}(i,j)} e^{\bar{k}(x_{i,j}, x_{p,q})}} \tag{6}$$

$$k(x_{i,j}, x_{m,n}) = -\frac{x_{i,j} - x_{m,n}}{\sigma_{i,j}^2} \tag{7}$$

where $\alpha_{i,j,m,n}$ denotes the pixel-level affinity function, $k$ and $\bar{k}$ represent the kernel function and average affinity value of $k$ over the image channels, respectively, and $\sigma_{i,j}$ denotes the standard deviation of the image intensity within the affinity kernel.

Pixel-adaptive mask refinement is a parameter-free recurrent method for local refinement with a computational speed much faster than that of GrabCut and dense CRFs. This makes the pixel-adaptive mask refinement suitable for embedding into the deep learning framework during training, not only a post-processing technique for segmentations. It is obvious that the number of iterations and affinity kernel size and shape are the main factors affecting the results based on Eq. (5). Furthermore, it has been proven that multiple kernels with different dilation rates often result in a better performance.

In this study, pixel-adaptive mask refinement is employed as a self-learning method to improve the segmentation performance of the desired model. With the refined semantic mask $o_{re}$, the corresponding pseudo-label $y_{re}$ can be obtained using a refined semantic mask ore. Given an unlabeled dataset containing $N$ images, self-learning (pixel-adaptive mask refinement) loss is defined as follows:

$$\mathcal{L}_{pamr}(X) = \mathcal{L}_{seg}(X, Y_{re}) = \frac{\sum_{l=1}^{N} 2 y_{re}^l o^l}{\sum_{l=1}^{N} y_{re}^l y_{re}^l + \sum_{l=1}^{N} o^l, o^l} \tag{8}$$

### 3.7. Overall loss function

In summary, we introduced multiple loss functions for the proposed source-free unsupervised domain adaptation framework. The framework is optimized by combining all loss functions in two stages. The overall loss function $\mathcal{L}_{all}$ is expressed as follows:

$$\mathcal{L}_{all} = \begin{cases} \lambda_1 \mathcal{L}_{fms}(X_t) + \lambda_2 \mathcal{L}_{ent}(X_t) + \lambda_3 \mathcal{L}_{seg}(X_t, Y_{1,t}) + \lambda_4 \mathcal{L}_{seg}(X_t, Y_{2,t}) & \text{if Epoch} < T \\ \lambda_1 \mathcal{L}_{fms}(X_t) + \lambda_2 \mathcal{L}_{ent}(X_t) + \lambda_3 \mathcal{L}_{seg}(X_t, Y_{1,t}) + \lambda_4 \mathcal{L}_{seg}(X_t, Y_{2,t}) & \text{if Epoch} \geq T \\ \quad + \lambda_5 \mathcal{L}_{pamr}(X_t) + \lambda_6 \mathcal{L}_{seg}(X_t, Y_{3,t}) & \end{cases} \quad (9)$$

where $\{\lambda_1, \lambda_2, \lambda_3, \lambda_4, \lambda_5, \lambda_6\}$ are the weights for adjusting the importance of each loss. Using the grid search method, they are set to $\{0.001, 10, 1, 1, 0.6, 0.3\}$ and remain constant throughout the training process. In the proposed framework, $T$ is recommended to be set sufficiently large to fully train the networks (U1-1, SC, U3) in the first stage; specifically, it is set as 150 in our task.

## 4. Experiment results and discussion

### 4.1. Dataset and metrics

In this study, we employed two public challenge datasets: the Multi-Atlas Labeling Beyond the Cranial Vault Challenge (Abdomen) containing 30 CT volumes released in MICCAI 2015 and CHAOS-Combined (CT-MR) Healthy Abdominal Organ Segmentation containing 20 T2-SPIR MR volumes released in ISBI 2019 (Kavur et al., 2019). Thirty volumes of CT images and 20 volumes of MR images were set as the source and target domains, respectively. The ground truth masks of both datasets included four organs (liver, right kidney, left kidney, and spleen), which were the objects to segment in our task. The MR volumes capture the abdominal area, whereas the CT volumes capture the area from the neck to the knee. We cropped the CT volumes to cover the abdominal organs, during which we aimed to segment and remove the background slices from each volume to retain target organ-containing slices. All 30 CT volumes were used to train the source model. The MR dataset was randomly divided into two parts; 16 volumes (80%) were used for training and four (20%) were used for testing. To be consistent with the CT dataset, we removed the background slices from the MR training dataset. The MR test dataset retained the background slices. All CT images were resized to a size of 256×256, which is the same as that of the MR images. The pixel values of the CT and MR images were within the ranges [-100, 400] and [0, 1200], respectively. The Dice similarity coefficient (DSC) and average symmetric surface distance (ASSD [voxel]) were used to evaluate the performance of the segmentation models.

### 4.2. Implementation details

All segmentation networks (U1 (U1-1 & U1-2), U2, and U3) adopted the classic U-Net structure (Figure 3) with different kernel numbers in the convolution blocks. U1-1 adopts the first convolution block of the classic U-Net, and U1-2 adopts the last eight blocks of the classic U-Net. The numbers of filters in convolution blocks in U1-1, U1-2, U2 and U3 are {64}, {128, 256, 512, 1024, 512, 256, 128, 64}, {64, 128, 256, 512, 1024, 512, 256, 128, 64}, and {16, 32, 64, 128, 256, 128, 64, 32, 16}, respectively. The style compensation network (SC), presented in Figure 6, contains seven stacked convolution layers, each followed by a batch normalization layer and an activation function layer (the first six layers are ReLU and last layer is sigmoid). The number of filters in the seven convolution layers are {64, 32, 16, 8, 4, 2, 1}, respectively. In the pixel-adaptive mask refinement module, the iterations, affinity kernel size, and composition of dilation rates are empirically set to 10, 3×3, [1, 2, 4, 8, 12, 24], respectively.

All experiments were implemented in Python 3.6.5, Pytorch 1.2.0 (Paszke et al., 2019) using two NVIDIA TITAN V GPUs, each with 12 GB of memory. The source model was trained on the source-labeled CT dataset in a fully supervised learning manner using the Adam optimizer (Kingma and Ba, 2014) (the learning rate, batch size, and epoch were set as 0.0001, 8, and 100, respectively). Segmentation networks U1 and U2 utilized the weights of the source model for initialization. In the proposed source-free unsupervised domain adaptation framework, the trainable part of U1, style compensation network (SC), and desired model U3 were trained with the RMSprop optimizer (Graves, 2013) (all smoothing constants were set to 0.9, the learning rates were set to 0.00012, 0.0004, and 0.0006, respectively; all batch sizes were set to 8). Note that the batch size of input of U1-1 should be the same as that of the source model during training to ensure the effectiveness of the feature map statistics loss.

### 4.3. Performance of the source-free unsupervised domain adaptation framework

Note that, for the proposed source-free unsupervised domain adaptation framework (Figure 2), each of the networks U1, U3, and the combination of U2 and SC, can be used for segmentation after training the framework. These networks are interdependent. Comparing their performances can help us further understand the proposed framework and desired network U3. Table 1 provides quantitative evaluations on the performances of U1, combination of U2 and SC (denoted as "U2"), and U3. We also included the results regarding the model trained and evaluated on the target MR images in a fully supervised learning manner (denoted as "Supervised learning"), which provides the performance upper bound. The results obtained directly from the model trained on the source CT images and evaluated on the target MR images (denoted as "W/o adaptation") provide the performance lower bound.

As reported in Table 1, U1, U2, and U3 achieved relatively good performances when segmenting abdominal organs from MR images with average Dice similarity coefficients of 0.844, 0.851, and 0.888, respectively, which are much better than the results without domain adaptation. Among them, the desired model U3 achieved the best performance in segmenting all four organs, particularly the spleen. Furthermore, the segmentations predicted by U3 were better than those predicted by U1 and U2 for all subjects in the testing dataset as shown in Figure 7.

Figure 8 shows example segmentation result images. Note that the source model pre-trained on the CT images cannot capture effective pixel information on abdominal organs from the MR images. The segmentation results from supervised learning yield the most accurate segmentation results. When using the proposed source-free unsupervised domain adaptation framework, the segmentations predicted by U1 and U2 either completely miss the structure of the spleen or mistake the nearby tissues as the spleen. Note that the segmentations predicted by U3 overcomes these problems and can depict more accurate organ structures, especially for the liver and its bright blood vessels (Figure 8).

Table 1 Quantitative evaluations on the performances of different networks in the proposed framework.

| Method | DSC ↑ | | | | | ASSD ↓ | | | | |
|---|---|---|---|---|---|---|---|---|---|---|
| | Liver | Kidney (R.) | Kidney (L.) | Spleen | Mean | Liver | Kidney (R.) | Kidney (L.) | Spleen | Mean |
| Supervised learning | 0.948 | 0.936 | 0.894 | 0.938 | 0.929 | 0.101 | 0.115 | 0.343 | 0.143 | 0.175 |
| W/o adaptation | 0.578 | 0.232 | 0.479 | 0.234 | 0.381 | 1.282 | 2.525 | 1.031 | 6.944 | 2.878 |
| U1 | 0.842 | 0.870 | 0.857 | 0.808 | 0.844 | 0.692 | 0.176 | 0.214 | 1.527 | 0.652 |
| U2 | 0.848 | 0.883 | 0.862 | 0.809 | 0.851 | 0.640 | 0.173 | **0.157** | 0.986 | 0.480 |
| U3 (desired) | **0.884** | **0.891** | **0.864** | **0.911** | **0.888** | **0.331** | **0.137** | 0.258 | **0.138** | **0.216** |

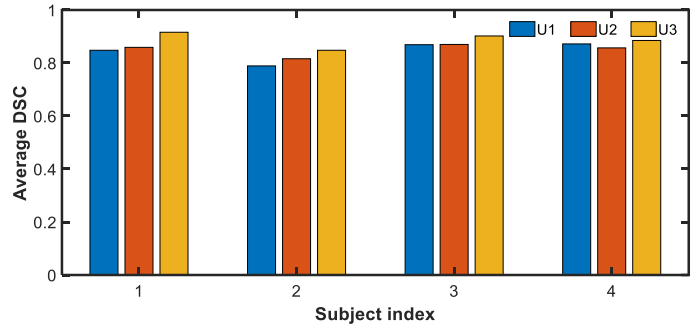

Figure 7 (Best viewed in color) Average dice similarity coefficient of all subjects in the testing dataset predicted by different networks in the proposed framework.

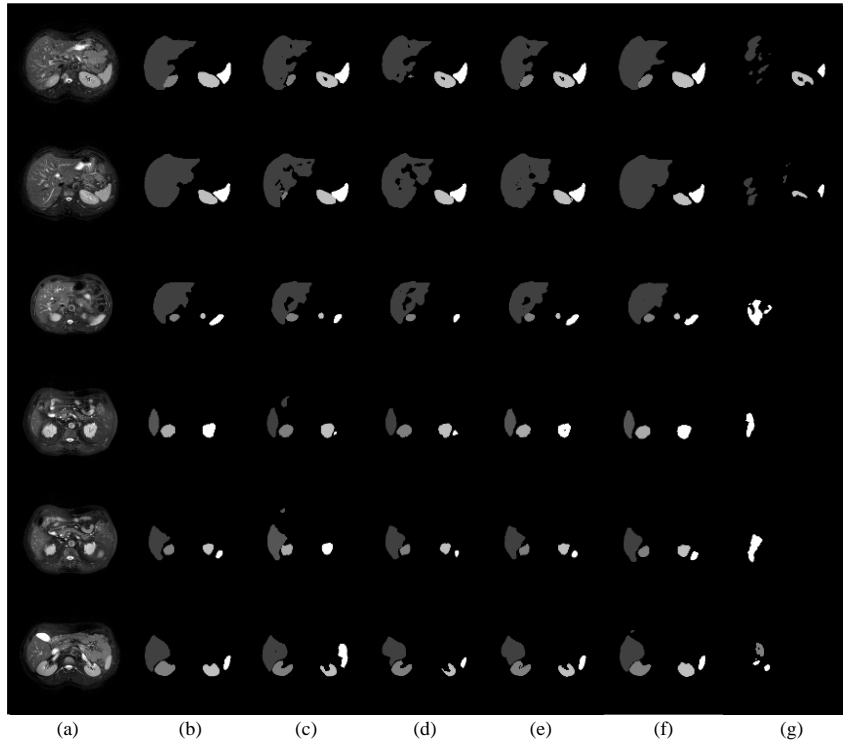

Figure 8 Examples of segmentation results predicted by different networks in the proposed framework. (a) MR image; (b) Ground truth; (c) Segmentation predicted by U1; (d) Segmentation predicted by U2; (e) Segmentation predicted by the desired model U3; (f) Segmentation under supervised learning; (e) Segmentation predicted by source model.

Figure 9 reports the average Dice similarity coefficient for various networks in the proposed framework during the training process. Note that the dice similarity coefficients of U2 and U3 exhibit the same trend during training. Both rise rapidly initially, slow down, and eventually enter a stable phase. Contrarily, the performance of U1 fluctuates until the 150$^{th}$ epoch. The performance suddenly increases at the 150$^{th}$ epoch and remains stable thereafter. Note that the performances of U2 and U3 exceed that of U1 after the 50$^{th}$ epoch. After the 150$^{th}$ epoch, the performances of U1 and U3 become comparable. The performance of the described network U3 is superior to that of both U1 and U3 after the 150$^{th}$ epoch. Here, we refer to the process before the 150$^{th}$ epoch as the first stage and that after the 150$^{th}$ epoch as the second stage.

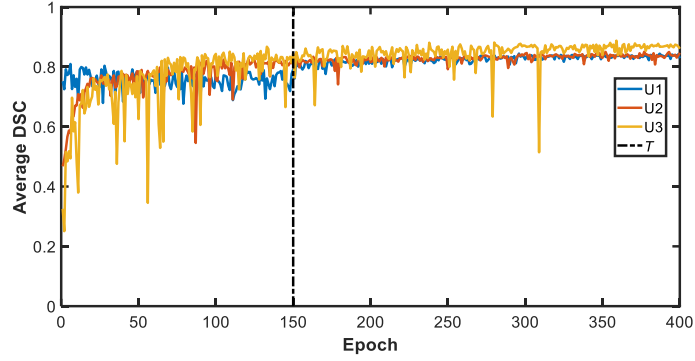

Figure 9 (Best viewed in color) Validation results of different networks in the proposed framework per epoch. *T* denotes the time (epoch) that the proposed framework enters the second stage.

At the beginning of the training process, the most useful information is the feature map statistics stored in model U1-2 that is pre-trained on the source CT images and combined with the entropy minimization loss to guide U1-1 to achieve domain adaptation. The performance of the combination of SC and U2 depends on U1 because the former is supervised by the latter. Similarly, the performance of U3 depends on that of U2. Therefore, the average dice similarity coefficient of U1 is higher than those of U2 and U3 before the 50$^{th}$ epoch. Subsequently, the style compensation network SC learns a better compensation coefficient to generate images similar to those present in the source domain, which improves the performance of U2 and U3. When the proposed framework enters the second stage (after the 150$^{th}$ epoch), pixel-adaptive mask refinement is added as a new guidance method to supervise the training of U3 to further improve its performance. The performance of U1 is also improved after adding the new guidance method provided by U3. The networks in the proposed framework form a circular learning mechanism in the second stage.

**4.4. Ablation study**

The proposed source-free unsupervised domain adaptation framework contains five key components: (i) feature map statistics-guided model adaptation, (ii) entropy minimization, (iii) style compensation, (iv) pixel-adaptive mask refinement, and (v) circular learning. In this section, we present detailed experiments to validate their effectiveness and analyze how these components contribute to the performance gains achieved by the proposed framework.

*4.4.1. The feature map statistics-guided model adaptation and entropy minimization*

As shown in Figure 2, the feature map statistics-guided model adaptation composed of U1-1 and U1-2 networks is used to segment target images; then, the entropy minimization loss is used as self-supervision to constrain the outputs from U1 to lower the entropy. Table 2 reports the performance of the desired model U3 when we remove the feature map statistics loss or entropy minimization loss. The former is denoted as "W/o FMS", and the latter as "W/o EMin".

Note that the proposed framework failed without the feature map statistics loss based on the Dice scores of all organs, even lower than the lower bound, as shown in Table 1. This is because U1 tends to output a segmentation with zero entropy (black image) when only entropy minimization loss is used. Under the "W/o EMin" setting, the segmentation performance of U3 decreases for all organs (the average Dice similarity coefficient decreases from 0.888 to 0.878). Figure 10 shows three examples of the entropy map predicted by U1 under the "W/o EMin" setting. Note that the entropy maps corresponding to "W/o

EMin" contain brighter pixels than those corresponding to "Proposed". These brighter pixels are mainly distributed inside the segmented organs, causing a decline in the segmentation accuracy. For example, the integrity of liver segmentation with "W/o EMin" is not as good as that with "Proposed" in the first and third rows. This means that the segmentation accuracy of U1 is relatively low, which reduces the segmentation accuracy of the desired model U3 to some extent, under the "W/o EMin" setting.

These results validate that both the feature map statistics-guided model adaptation and entropy minimization are important components in the proposed source-free unsupervised domain adaptation framework, with the former playing a more crucial role in domain adaptation.

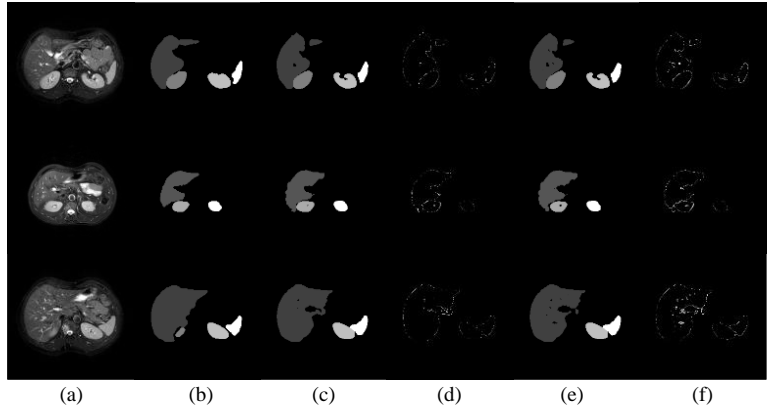

(a)　　　　(b)　　　　(c)　　　　(d)　　　　(e)　　　　(f)

Figure 10 Examples of entropy map predicted by U1 under the "W/o EMin" setting. (a) MR image; (b) Ground truth; (c) Segmentation under the "Proposed" setting; (d) Entropy map under the "Proposed" setting; (e) Segmentation under the "W/o EMin" setting; (e) Entropy map under the "W/o EMin" setting.

Table 2 Quantitative evaluations on the performances of the desired model U3 under different ablation study settings.

| Method | DSC ↑ | | | | | ASSD ↓ | | | | |
|---|---|---|---|---|---|---|---|---|---|---|
| | Liver | Kidney (R.) | Kidney (L.) | Spleen | Mean | Liver | Kidney (R.) | Kidney (L.) | Spleen | Mean |
| Proposed | **0.884** | **0.891** | **0.864** | **0.911** | **0.888** | **0.331** | **0.137** | 0.258 | **0.138** | **0.216** |
| W/o FMS | 0.000 | 0.000 | 0.000 | 0.000 | 0.000 | 9999 | 9999 | 9999 | 9999 | 9999 |
| W/o EMin | 0.868 | 0.886 | 0.857 | 0.903 | 0.878 | 0.380 | 0.147 | 0.263 | 0.253 | 0.258 |
| W/o SC | 0.876 | 0.880 | 0.830 | 0.830 | 0.854 | 0.426 | 0.211 | 0.192 | 1.033 | 0.465 |
| With ST | 0.862 | 0.885 | 0.863 | 0.867 | 0.870 | 0.598 | 0.148 | 0.194 | 0.610 | 0.417 |
| W/o PAMR | 0.872 | 0.889 | 0.849 | 0.841 | 0.863 | 0.349 | 0.138 | 0.450 | 0.628 | 0.391 |
| W/o CL | 0.875 | 0.878 | 0.863 | 0.882 | 0.875 | 0.427 | 0.141 | **0.168** | 0.241 | 0.244 |

*4.4.2. The style compensation*

In our proposed framework, we designed a style compensation network (SC) to output the compensation coefficient $p_s$ for generating source-like images $x_{t \to s}$ guided by the pseudo-label $y_{1,t}$. The style compensation network combined with the pre-trained source model U2 is used to generate guidance $y_{2,t}$ for the desired model U3. Compared with $y_{1,t}$, guidance $y_{2,t}$ is more accurate for training model U3. This is demonstrated in Figure 9. However, it is still necessary to investigate whether the performance of U3 declines without style compensation. Thus, we performed an ablation study by removing the combination of SC and U2 and letting $y_{1,t}$ guide U3 directly (denoted as "W/o SC").

Table 2 and Figure 11 highlight the performance of U3 under the "W/o SC" setting. Table 2 shows that the performance of U3 is significantly decreased regarding the segmentation of all organs. This trend is especially observable for the spleen and left kidney, with DSCs decreasing from 0.911 to 0.830 and from 0.864 to 0.830, respectively. Figure 11 shows that the segmentation accuracy of all subjects without the guidance of $y_{2,t}$ is lower than that with it. Figure 12 shows several examples of segmentation results predicted by U3 under the "W/o SC" setting. Note that U3 tends to classify the unrelated tissues around the spleen as spleen tissues (the first and second rows in Figure 12) or lose the structures of the entire (the third row) or part (the fourth row) of the left kidney. These results reveal that, compared to U1, the combination of the style compensation network and U2 can provide more accurate guidance for training the desired model U3.

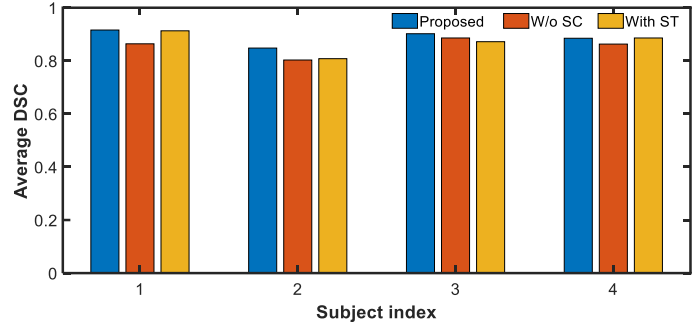

Figure 11 (Best viewed in color) Average dice similarity coefficient of all subjects in testing dataset under the "W/o SC" and "With ST" settings.

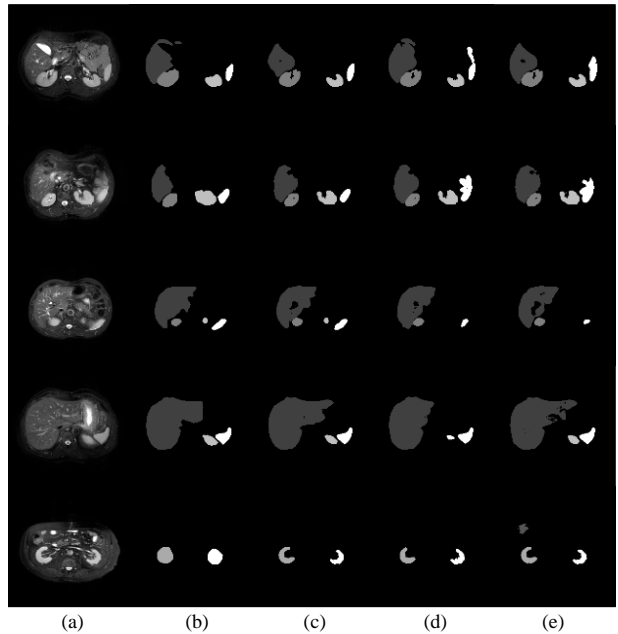

Figure 12 Examples of segmentation results predicted by U3 under the "W/o SC" and "With ST" settings. (a) MR image; (b) Ground truth; (c) Segmentation under the "Proposed" setting; (d) Segmentation under the "W/o SC" setting; (e) Segmentation under the "With ST" setting.

As shown in Figure 2, the target image $x_t$ is multiplied by the compensation coefficient $p_s$ to obtain the source-like image $x_{t\rightarrow s}$. We conducted experiments using the same network structures without multiplying the target image $x_t$ by the compensation coefficient $p_s$. The network SC works as a style transformation (ST) network that outputs the source-like image $x_{t\rightarrow s}$ directly, with the target image $x_t$ being the input in this case. We denote this experiment as "With ST". Table 2 and Figure 11 report the performance result for U3 in this case. Note that the performance of U3 decreases for all organs (especially for the spleen and liver where the Dice similarity coefficient decreases from 0.911 to 0.867 and from 0.884 to 0.862, respectively) and subjects. Figure 12 illustrates several example slices. Note that under the case "With ST", U3 tends to yield wrong predictions of spleen tissues (the first and second rows of Figure 12) or lose the structure of the entire (the third row) or part (the fourth row) of the left kidney. However, the loss degree is less than that observed with the "W/o SC" setting. Moreover, U3 occasionally misjudges the abdominal skin as the liver (fifth row).

Figure 13 illustrates two example slices of style transformation images (the input of U2) and the corresponding segmentations (the output of U2) under the "With ST" setting. Note that U2 suffers from the same misjudgment as U3, shown in Figure 12e. Because U2 supervises the learning of U3, the prediction error from U2 can lead to the prediction errors from U3. The prediction errors from U2 occur because the pixel values of intestine and spleen (abdominal skin and liver) tend to be the same in the transformed images obtained under the "With ST" setting, while the transformed images obtained with style compensation method can better preserve the contrast between these organs in the original MR images.

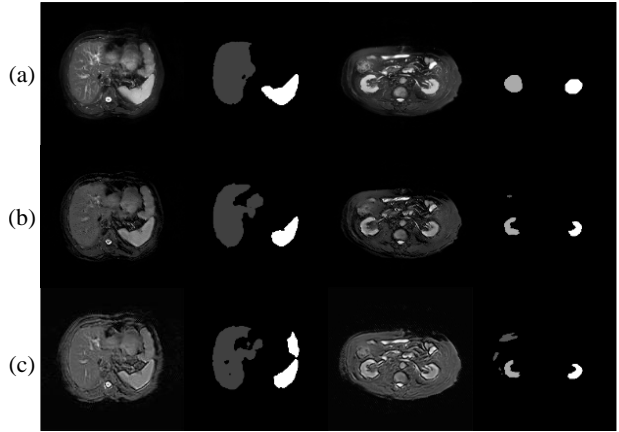

Figure 13 Examples of style transformation images and segmentation results predicted by U2 under the "Proposed" and "With ST" settings. (a) Original MR image and ground truth; (b) Transformed image $x_{t \to s}$ and segmentation under the "Proposed" setting; (c) Transformed image $x_{t \to s}$ and segmentation under the "With ST" setting.

*4.4.3. The pixel-adaptive mask refinement and circular learning*

In the proposed domain adaptation framework, pixel-adaptive mask refinement is used for self-learning to further improve the performance of U3. To validate the effectiveness of this method, Table 2 reports the comparison between the performance of U3 with (denoted as "Proposed") and without (denoted as "W/o PAMR") pixel-adaptive mask refinement. Note that the Dice similarity coefficient drops significantly in all organs under the "W/o PAMR" setting. Especially in the spleen and left kidney, the Dice similarity coefficient drops from 0.911 to 0.841 and from 0.864 to 0.849, respectively. Figure 14 shows three representative example slices of the segmentations obtained by U3 with and without the pixel-adaptive mask refinement method. Note that the proposed method can capture more complete structures of the liver (the first and third rows of Figure 13), spleen (the first row), and left kidney (the second row) and can eliminate unreasonable isolated pixels (the third row). These improvements are attributed to the fact that pixel-adaptive mask refinement itself is a method for improving the local pixel consistency of segmentations.

These observations confirm the positive effect of pixel-adaptive mask refinement when it is used as a self-learning method on the performance of U3. Note that the two-stage process proposed in this study is necessary to ensure that the segmentation performance is improved by using pixel-adaptive mask refinement. We include pixel-adaptive mask refinement in U3 during the second stage because the segmentations can capture the structure of the abdominal organs well. If the pixel-adaptive mask refinement is used during the initial phase of the training when the segmentations are poor, the refinement loss will mislead the learning of U3 and may result in failing to train U3.

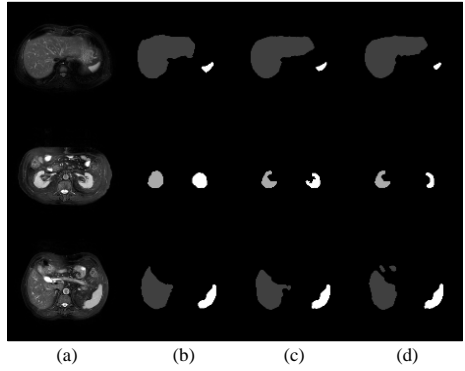

(a)      (b)      (c)      (d)

Figure 14 Examples of segmentation results predicted by U3 under the setting of "W/o PAMR". (a) MR image; (b) Ground truth; (c) Segmentation under the setting of "Proposed"; (d) Segmentation under the setting of "W/o PAMR".

During the second stage, the pseudo-label $y_{3,t}$ generated by U3 is added to supervise U1-1, imposing the so-called circular learning methodology on the entire framework as shown in Figure 2. Table 2 reports the performance variations in U3 if we remove the supervision of $y_{3,t}$ on U1-1 from the framework. Note that performance decreased regarding all organs. As shown in Figure 9, the performance of U3 exceeds that of U1 at the end of the first stage. Circular learning method can result in transferring richer semantic information from U3 to U1. This positive feedback optimization system further improves the final performance of the desired model U3. Similar to pixel-adaptive mask refinement, circular learning will also lead to a failure in training U3 if we introduce this component at the beginning of the training.

### 4.5. Comparison with the state-of-the-art methods

In this section, we compare our proposed source-free unsupervised domain adaptation method with the state-of-the-art source-free and source-involved domain adaptation approaches (Bateson et al., 2021; Chen et al., 2019a; Chen et al., 2020; Chen et al., 2021; Hoffman et al., 2018; Huo et al., 2018; Tsai et al., 2018; Wang et al., 2020; Zhu et al., 2017). Tent (Wang et al., 2020), DPL (Chen et al., 2021) and SFDA (Bateson et al., 2021) are source-free domain adaptation methods, whereas SIFA_v2 (Chen et al., 2020), SIFA_v1 (Chen et al., 2019a), SynSeg-Net (Huo et al., 2018), AdaOutput (Tsai et al., 2018), CycleGAN (Zhu et al., 2017) and CyCADA (Hoffman et al., 2018) are source-involved domain adaptation methods. In source-free methods, the SFDA requires prior class-ratio information of unlabeled target data derived from approximate anatomical knowledge during domain adaptation. For ease of comparison, we directly provide the ground truth class-ratio for supervising the target model as the upper bound of the SFDA. Table 3 presents the corresponding comparison results. Note that the performances of the source-involved unsupervised domain adaptation methods are quoted directly from the reference (Chen et al., 2020) because the data used in this study, performance evaluation method, and task to be solved are the same as those considered in the reference.

As Table 3 shows, our proposed method achieves the best performance with an average Dice similarity coefficient of 0.888. Compared with other domain adaptation methods, our approach is more balanced with respect to the segmentation performance of each organ and excels in the segmentation of the left kidney and spleen. Among the source-free domain adaptation methods, SFDA with a prior ground truth class-ratio outperforms Tent and DPL. The prior ground truth class-ratio information of the target data is a strong constraint, resulting in a high performing SFDA model. However, obtaining the ground truth class-ratio of the unlabeled target data is costly. The authors (Bateson et al., 2021) proposed utilizing the estimated class-ratio instead of the ground truth class-ratio using anatomical knowledge to reduce the cost. In source-involved unsupervised domain adaptation, the three methods (SIFA_v2, SIFA_v1, and CyCADA) of simultaneously aligning features and images achieved the top three performances, demonstrating obvious advantages compared with those aligning only images or features. Note that our proposed framework leverages three crucial methods (reviewed in section 2.1) for domain adaptation: aligning features (in the top segmentation network U1), aligning images (in the combination of the style compensation network SC and the segmentation network U2), and self-learning (pixel-adaptive mask refinement and circular learning). This is likely the reason why our proposed source-free unsupervised domain adaptation framework can achieve the best performance, even without accessing the source dataset.

Table 3 Performance comparison between the proposed framework and other domain adaptation methods (adaptation from CT to MR).

| Method | Source free | Supervision | DSC ↑ | | | | |
|---|---|---|---|---|---|---|---|
| | | | Liver | Kidney (R.) | Kidney (L.) | Spleen | Mean |
| W/o adaptation (lower bound) | Yes | None | 0.578 | 0.232 | 0.479 | 0.234 | 0.381 |
| Tent (Wang et al., 2020) | Yes | None | 0.671 | 0.805 | 0.739 | 0.414 | 0.657 |
| DPL (Chen et al., 2021) | Yes | None | 0.790 | 0.811 | 0.742 | 0.638 | 0.745 |
| SFDA (Bateson et al., 2021) | Yes | Ground truth class-ratio | 0.885 | 0.876 | 0.823 | 0.889 | 0.868 |
| Proposed | Yes | None | 0.884 | 0.891 | **0.864** | **0.911** | **0.888** |
| SIFA_v2 (Chen et al., 2020) | No | None | **0.900** | 0.891 | 0.802 | 0.823 | 0.854 |
| SIFA_v1 (Chen et al., 2019a) | No | None | 0.885 | 0.900 | 0.797 | 0.813 | 0.849 |
| SynSeg-Net (Huo et al., 2018) | No | None | 0.872 | **0.902** | 0.766 | 0.796 | 0.834 |
| AdaOutput (Tsai et al., 2018) | No | None | 0.858 | 0.897 | 0.763 | 0.822 | 0.835 |
| CycleGAN (Zhu et al., 2017) | No | None | 0.888 | 0.873 | 0.768 | 0.794 | 0.831 |
| CyCADA (Hoffman et al., 2018) | No | None | 0.887 | 0.893 | 0.781 | 0.802 | 0.841 |

**4.6. Extension of the proposed framework with one target labeled volume**

Source-free unsupervised domain adaptation is used to avoid heavy annotation requirements in the target domain. However, minor annotation costs are incurred, which are generally affordable. In this section, we describe an extension module that can be easily embedded into the original proposed framework so that the performance of the proposed method can be further improves when annotation data (labeled MR images) exist. The extension module is presented in Figure 2. With this approach, after every training epoch of U3 in the original proposed framework, the labeled MR images are used to train U3 for one additional epoch in a fully supervised learning manner. To demonstrate this method, one subject is randomly selected from the training dataset, and the corresponding single-labeled volume is used. The epoch, optimizer, batch size, and learning rate are the same as those used in unsupervised domain adaptation training.

Table 4 reports the differences in the performance of U3 after the extension module is integrated into the proposed framework, denoted as "Proposed with extension", by using only one labeled MR volume. Note that after adding only one labeled volume, the performance of the proposed method significantly improves regarding all organs, and the average Dice similarity coefficient increases from 0.888 to 0.922, which is almost equal to the upper bound (0.929).

Table 4 Quantitative evaluation on the performance of the desired model U3 in the proposed framework with an extension module.

| Method | DSC ↑ | | | | | ASSD ↓ | | | | |
|---|---|---|---|---|---|---|---|---|---|---|
| | Liver | Kidney (R.) | Kidney (L.) | Spleen | Mean | Liver | Kidney (R.) | Kidney (L.) | Spleen | Mean |
| Supervised learning | 0.948 | 0.936 | 0.894 | 0.938 | 0.929 | 0.101 | 0.115 | 0.343 | 0.143 | 0.175 |
| Proposed | 0.884 | 0.891 | 0.864 | 0.911 | 0.888 | 0.331 | 0.137 | 0.258 | 0.138 | 0.216 |
| Proposed with extension | 0.926 | 0.918 | 0.916 | 0.930 | 0.922 | 0.202 | 0.121 | 0.126 | 0.159 | 0.152 |

**4.7. Cross-modality adaptation in reverse direction**

In this section, we demonstrate that the proposed method is suitable for domain adaptation in the reverse direction (from the labeled MR dataset to the unlabeled CT dataset). All 20 MR volumes after removing the background slices were used to train the source model in a supervised learning manner. The CT dataset without background slices was randomly divided into two parts with 24 volumes (80%) for training and 6 volumes (20%) for testing. The structure of all network models and parameter settings required for training were the same as those used in the domain adaptation from CT to MR, except for the structure of U1-1. Note that the structure of U1-1 adopted the first convolution block of the classic U-Net, which contains two convolution layers with the same parameters when implementing domain adaptation from CT to MR. In the adaptation from MR to CT, the structure of U1-1 adopted five convolution layers with the same parameters in our experiment using the grid search method.

Table 5 presents the experimental results obtained using the proposed and nine other domain adaptation methods. Note that the proposed method still achieved good performance with Dice similarity coefficients of 0.881, 0.808, 0.881, and 0.792 for the liver, right kidney, left kidney, and spleen, respectively. The achieved performance was second only to that of SFDA, which was supervised by the ground truth class-ratio. Comparing Table 5 with Table 3, the overall performance of the adaptation from MR to CT was worse than that from CT to MR. Note that the performance of DPL in adaptation from MR to CT significantly degrades compared with that from CT to MR. We speculate that the predictions of CT dataset using the source model trained with the MR dataset may be highly noisy, resulting in the DPL method failing, which depends on the noise level of the pseudo-label.

Table 5 Performance comparison between the proposed framework and other domain adaptation methods (adaptation from MR to CT)

| Method | Source free | Supervision | DSC ↑ | | | | |
|---|---|---|---|---|---|---|---|
| | | | Liver | Kidney (R.) | Kidney (L.) | Spleen | Mean |
| W/o adaptation (lower bound) | Yes | None | 0.385 | 0.166 | 0.182 | 0.463 | 0.299 |
| Tent (Wang et al., 2020) | Yes | None | 0.757 | 0.753 | 0.596 | 0.503 | 0.652 |
| DPL (Chen et al., 2021) | Yes | None | 0.683 | 0.074 | 0.115 | 0.000 | 0.218 |
| SFDA (Bateson et al., 2021) | Yes | Ground truth class-ratio | 0.877 | **0.844** | 0.865 | **0.872** | **0.864** |
| Proposed | Yes | None | **0.881** | 0.808 | **0.881** | 0.792 | 0.841 |
| SIFA_v2 (Chen et al., 2020) | No | None | 0.880 | 0.833 | 0.809 | 0.826 | 0.837 |
| SIFA_v1 (Chen et al., 2019a) | No | None | 0.879 | 0.837 | 0.801 | 0.805 | 0.831 |
| SynSeg-Net (Huo et al., 2018) | No | None | 0.850 | 0.821 | 0.727 | 0.810 | 0.802 |
| AdaOutput (Tsai et al., 2018) | No | None | 0.854 | 0.797 | 0.797 | 0.817 | 0.816 |
| CycleGAN (Zhu et al., 2017) | No | None | 0.834 | 0.793 | 0.794 | 0.773 | 0.799 |
| CyCADA (Hoffman et al., 2018) | No | None | 0.845 | 0.786 | 0.803 | 0.769 | 0.801 |

**5. Conclusion**

In this study, we propose an effective source-free unsupervised domain adaptation framework for cross-modality abdominal multi-organ segmentation from a source-labeled CT dataset to target an unlabeled MR dataset. The proposed method can be used to avoid high annotation costs in the target domain and protect data privacy in the source domain. Our proposed framework includes five key components in a two-stage process. The framework is optimized in a one-way manner during the first stage and in a circular manner during the second stage. Extensive experiments are conducted to validate the effectiveness of the key components. Using the proposed framework, we achieved high segmentation performance regarding the organs of interest, namely, the liver, right kidney, left kidney, and spleen, with dice similarity coefficients of 0.884, 0.891, 0.864, and 0.911, respectively, outperforming nine state-of-the-art domain adaptation methods. We explored the scalability of the proposed framework in the presence of minimal labeled data in the target domain. Our results show that, by adding a simple extension module, the proposed method can achieve an average Dice similarity coefficient of 0.922 using a single labeled volume, which is close to the upper bound (0.929) determined using fully supervised learning. In addition, we demonstrated the effectiveness of our method for source-free unsupervised domain adaptation in the reverse direction as well.

**Declaration of competing interest**

The authors declare that there are no conflicts of interest regarding the publication of this paper.

**CRediT authorship contribution statement**

**Jin Hong**: Conceptualization; Data curation; Formal analysis; Investigation; Methodology; Software; Validation; Visualization; Writing-original draft. **Yu-Dong Zhang**: Investigation; Validation; Resources; Supervision; Writing-review & editing. **Weitian Chen**: Funding acquisition; Project administration; Resources; Supervision; Writing-review & editing.

**Acknowledgements**

This study was supported by a Faculty Innovation Award from the Faculty of Medicine of The Chinese University of Hong Kong, and a grant from the Innovation and Technology Commission of the Hong Kong SAR (Project MRP/046/20X).

Segmentation Challenge Data, v1.03 ed. Zenodo.

Kim, Y., Cho, D., Han, K., Panda, P., Hong, S., 2021. Domain adaptation without source data. IEEE Transactions on Artificial Intelligence.

Kingma, D.P., Ba, J., 2014. Adam: A method for stochastic optimization. arXiv preprint arXiv:1412.6980.

Kouw, W.M., Loog, M., 2019. A review of domain adaptation without target labels. IEEE transactions on pattern analysis and machine intelligence 43, 766-785.

Krähenbühl, P., Koltun, V., 2011. Efficient inference in fully connected crfs with gaussian edge potentials. Advances in neural information processing systems 24, 109-117.

Kundu, J.N., Venkat, N., Babu, R.V., 2020. Universal source-free domain adaptation, Proceedings of the IEEE/CVF Conference on Computer Vision and Pattern Recognition, pp. 4544-4553.

Laine, S., Aila, T., 2016. Temporal ensembling for semi-supervised learning. arXiv preprint arXiv:1610.02242.

Li, R., Jiao, Q., Cao, W., Wong, H.-S., Wu, S., 2020. Model adaptation: Unsupervised domain adaptation without source data, Proceedings of the IEEE/CVF Conference on Computer Vision and Pattern Recognition, pp. 9641-9650.

Liang, J., He, R., Sun, Z., Tan, T., 2019. Distant supervised centroid shift: A simple and efficient approach to visual domain adaptation, Proceedings of the IEEE/CVF Conference on Computer Vision and Pattern Recognition, pp. 2975-2984.

Liang, J., Hu, D., Feng, J., 2020. Do we really need to access the source data? source hypothesis transfer for unsupervised domain adaptation, International Conference on Machine Learning. PMLR, pp. 6028-6039.

Liang, J., Hu, D., Wang, Y., He, R., Feng, J., 2021. Source data-absent unsupervised domain adaptation through hypothesis transfer and labeling transfer. IEEE Transactions on Pattern Analysis and Machine Intelligence.

Liu, Y., Zhang, W., Wang, J., 2021. Source-free domain adaptation for semantic segmentation, Proceedings of the IEEE/CVF Conference on Computer Vision and Pattern Recognition, pp. 1215-1224.

Long, M., Zhu, H., Wang, J., Jordan, M.I., 2016. Unsupervised domain adaptation with residual transfer networks. arXiv preprint arXiv:1602.04433.

Luo, X., Chen, J., Song, T., Wang, G., 2020. Semi-supervised medical image segmentation through dual-task consistency. arXiv preprint arXiv:2009.04448.

Luo, X., Liao, W., Chen, J., Song, T., Chen, Y., Zhang, S., Chen, N., Wang, G., Zhang, S., 2021. Efficient semi-supervised gross target volume of nasopharyngeal carcinoma segmentation via uncertainty rectified pyramid consistency, International Conference on Medical Image Computing and Computer-Assisted Intervention. Springer, pp. 318-329.

Mathieu, M., Zhao, J., Sprechmann, P., Ramesh, A., LeCun, Y., 2016. Disentangling factors of variation in deep representations using adversarial training. arXiv preprint arXiv:1611.03383.

Milletari, F., Navab, N., Ahmadi, S.-A., 2016. V-net: Fully convolutional neural networks for volumetric medical image segmentation, 2016 fourth international conference on 3D vision (3DV). IEEE, pp. 565-571.

Nair, V., Hinton, G.E., 2010. Rectified linear units improve restricted boltzmann machines, Icml.

Novosad, P., Fonov, V., Collins, D.L., 2019. Unsupervised domain adaptation for the automated segmentation of neuroanatomy in MRI: a deep learning approach. bioRxiv, 845537.

Pan, S.J., Yang, Q., 2009. A survey on transfer learning. IEEE Transactions on knowledge and data engineering 22, 1345-1359.

Paszke, A., Gross, S., Massa, F., Lerer, A., Bradbury, J., Chanan, G., Killeen, T., Lin, Z., Gimelshein, N., Antiga, L., 2019. Pytorch: An imperative style, high-performance deep learning library. arXiv preprint arXiv:1912.01703.

Perone, C.S., Ballester, P., Barros, R.C., Cohen-Adad, J., 2019. Unsupervised domain adaptation for medical imaging segmentation with self-ensembling. NeuroImage 194, 1-11.

Ronneberger, O., Fischer, P., Brox, T., 2015. U-net: Convolutional networks for biomedical image segmentation, International Conference on Medical image computing and computer-assisted intervention. Springer, pp. 234-241.

Rother, C., Kolmogorov, V., Blake, A., 2004. " GrabCut" interactive foreground extraction using iterated graph cuts. ACM transactions on graphics (TOG) 23, 309-314.